\documentclass[letterpaper]{article}
\usepackage[preprint]{aaai2027}
\usepackage[hyphens]{url}
\usepackage{graphicx}
\usepackage{natbib}
\usepackage{caption}
\usepackage{booktabs}       
\usepackage{amsmath}        

\title{Event-Causal RAG: A Retrieval-Augmented Generation Framework for Long Video Reasoning in Complex Scenarios}

\author{
Peizheng Yan\textsuperscript{\rm 1},
Yu Zhao\textsuperscript{\rm 2}\thanks{Corresponding authors.},
Liang Xie\textsuperscript{\rm 3,4}\footnotemark[1],\\
Juntong Qi\textsuperscript{\rm 5},
Mingming Wang\textsuperscript{\rm 1},
Erwei Yin\textsuperscript{\rm 3,4}
}
\affiliations{
\textsuperscript{\rm 1}Tianjin Key Lab of Intelligent Unmanned Swarm Tech \& System,
Tianjin University, Tianjin, China\\
\textsuperscript{\rm 2}Institute of Computing and Intelligence,
Harbin Institute of Technology (Shenzhen), Shenzhen, China\\
\textsuperscript{\rm 3}Tianjin Artificial Intelligence Innovation Center, Tianjin, China\\
\textsuperscript{\rm 4}Defense Innovation Institute, Academy of Military Sciences, Beijing, China\\
\textsuperscript{\rm 5}School of Future Technology, Shanghai University, Shanghai, China\\
\texttt{ypzyh@tju.edu.cn}
}

\begin{document}

\maketitle

\begin{abstract}
Large vision-language models perform well on short- and medium-length video understanding but still struggle to maintain coherent event memory and recover long-range relationships in ultra-long videos. End-to-end methods are limited by visual-token growth and context length, while fixed-segment retrieval often fragments complete events and weakens state-transition modeling.
We propose Event-Causal RAG (EC-RAG), a lightweight retrieval-augmented framework for ultra-long and streaming video reasoning. A dual visual-audio sentinel mechanism segments video streams into semantically complete events, represented as State-Event-State (SES) structures that organize observable pre-event states, central events, and post-event states as event-local causal transitions. These transitions are stored in dual vector-graph memory and temporally connected through entity-consistent trajectories. During question answering, bidirectional graph retrieval recovers relevant predecessor and successor events, and answers are generated using both structured memory and the corresponding video evidence.
We further introduce ECV-1H, an hour-scale long-video QA benchmark dedicated to directed event-causal reasoning, with all source videos exceeding one hour. It covers over 150 hours of untrimmed video and contains 1,251 fully human-annotated QA pairs. 
EC-RAG improves overall accuracy by 4.96\%--11.67\% across three open-source video foundation models and achieves consistent gains across public datasets. 
On a single RTX 5090 GPU with 32 GB of memory, EC-RAG can continuously process videos while maintaining controlled streaming memory usage.
\end{abstract}

\section{Introduction}
\label{sec:intro}
Recent large vision-language models (LVLMs) have substantially improved short- and medium-length video understanding \cite{li2024llavanext,he2024malmm,ranasinghe2024understanding,zhang2024survey,wang2025videoxl,huang2024longvlm}.
These gains, however, do not directly transfer to ultra-long or continuously arriving streams, which require tracking evolving storylines, maintaining long-term memory, and inferring causal dependencies across minutes or hours \cite{mangalam2024egoschema,liu2024longvideobench,fu2024videomme,du2024eventoriented,liu2024causality}. These capabilities are central to surveillance~\cite{wang2025eventvad,ali2024bilstm}, medical monitoring~\cite{sathyanarayana2018vision}, and traffic analysis~\cite{s22176563,pawar2022roadaccidents}.

\begin{figure}[t]
    \centering
    \includegraphics[width=\columnwidth]{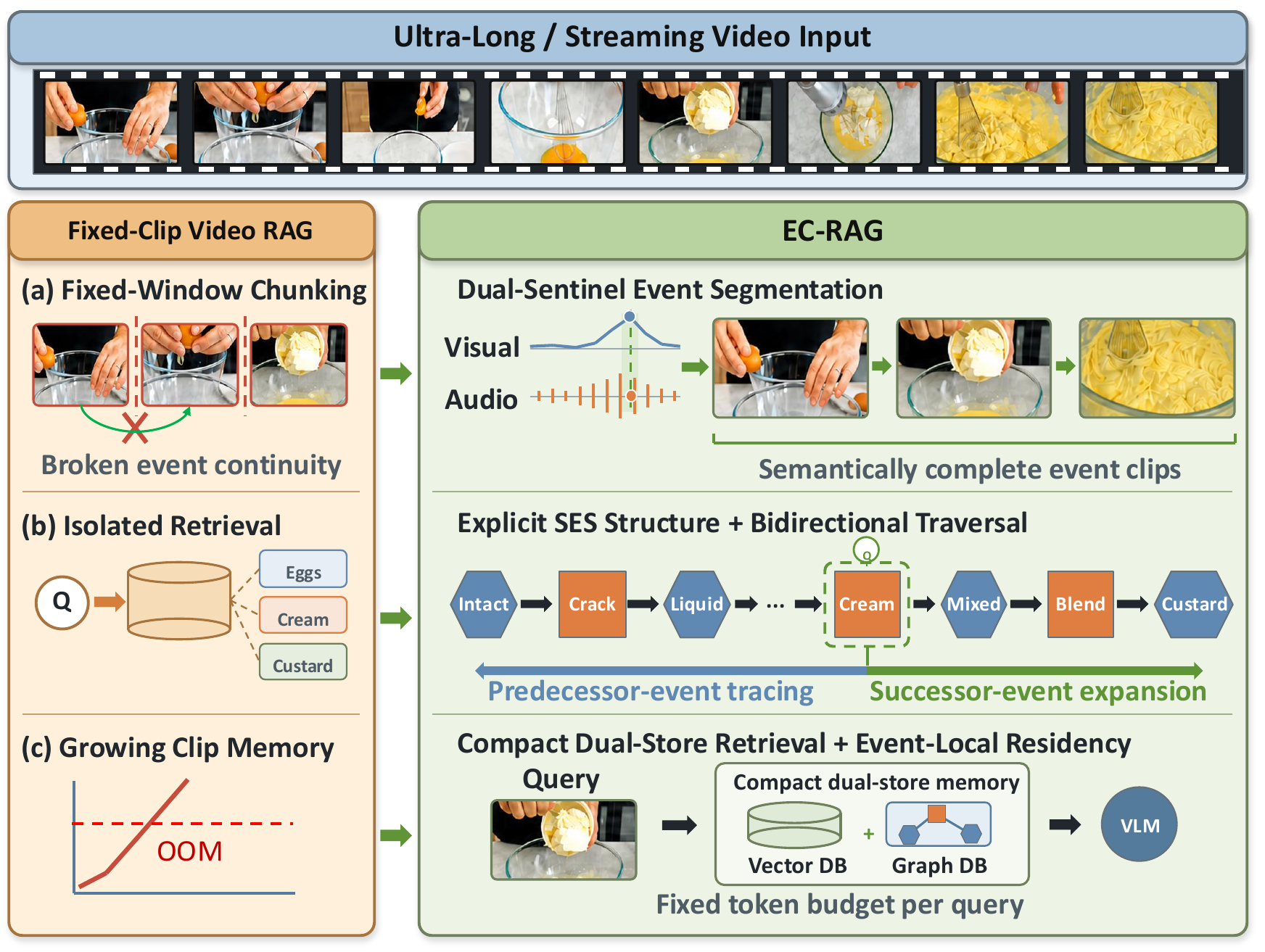}
    \caption{Comparison of conventional video RAG and EC-RAG.}
    \label{fig:teaser}
\end{figure}

Current end-to-end VLM-based video understanding solutions face a severe computational bottleneck.
As video duration grows, the tokenized representation quickly becomes prohibitively large, leading to excessive latency, memory overflow, and unstable inference \cite{souza2024temporal, chen2024mecd, liu2024causality}. 
Although recent efforts have explored temporal pooling, token compression, sparse attention, and context window extension to alleviate this issue \cite{wei2024visual, ge2025v2pe, ge2024adaptive, zhao2024minicache, shin2025infinipot, dai2024corm, liu2024pyramid, zheng2024cachegen}, these techniques do not fundamentally resolve the challenge of reasoning over extremely long sequences. 
In practice, models still suffer from the well-known lost-in-the-middle phenomenon \cite{liu2024lost}, as well as rising computational costs as context windows grow.

To avoid processing an entire long video at once, recent efforts have adopted retrieval-augmented generation (RAG) for long-video understanding \cite{jeong2025videorag,luo2024video,VideoRAG}.
Nevertheless, existing video-RAG approaches still exhibit three critical limitations:
\textbf{(1)} They often \textbf{fragment spatiotemporal context} by constructing memory over fixed-duration clips rather than semantically complete events. 
In most existing pipelines, a long video is mechanically segmented into windows of equal length, e.g., every 5 or 10 seconds, and each segment is independently embedded.
While convenient, this design ignores the natural boundaries of events. 
A coherent event and its associated state transition may therefore be split across multiple clips \cite{du2024eventoriented, rodin2024action, chen2024mecd}.
\textbf{(2)} Existing retrieval strategies primarily rely on \textbf{global semantic similarity} between the query and individual clips, while largely ignoring event-induced state changes and their long-range temporal dependencies. 
This design is sufficient when the goal is to find a visually similar scene or a directly matching clip, but it is inadequate for questions that require multi-step reasoning across distant observations \cite{xiao2021next, mangalam2024egoschema, zhang2024multimodal, li2024causaltad}.
\textbf{(3)} Current RAG-based methods \textbf{remain costly in both storage and online processing}. 
Because they store dense representations for large numbers of low-level clips,
the memory overhead grows rapidly with video duration.
This issue becomes more severe in streaming environments, where the system must continuously process and retrieve new content as the video arrives \cite{chen2024streaming, anonymous2024streamingvlm, yan2025streaming, chatterjee2025streamingvllm, zhao2025streamchat}. 
Meanwhile, maintaining near-real-time large-model inference over high-throughput streams remains computationally demanding~\cite{anonymous2024streamingvlm,chatterjee2025streamingvllm}.

As illustrated in Figure~\ref{fig:teaser}, we propose Event-Causal RAG, a retrieval-augmented framework for reasoning over ultra-long and streaming video. By releasing completed visual states at event boundaries, EC-RAG prevents visual residency from accumulating with stream duration, enabling continuous 24-hour-plus processing with negligible incremental storage pressure.
Our key idea is to replace clip-level memory with a State-Event-State (SES) graph memory that aligns retrieval units with semantically complete events rather than arbitrary temporal windows. 
Specifically, we first perform asynchronous event-driven segmentation on streaming videos using joint visual and audio cues, so that each memory unit preserves a naturally coherent event whenever possible. 
Next, we parse each clip into an SES record, where the central event connects the observable states of its participating entities before and after the event, forming an event-local causal transition.
These records are aligned through entity coreference and organized into a global event-state graph stored in dual memory: a Vector DB for semantic recall and a Graph DB for spatiotemporal topology.
Compared with conventional vector databases that mainly support similarity retrieval, this memory also traces the successive states of the same entity and the events associated with each state transition.
Finally, bidirectional graph retrieval collects relevant event-local SES transitions together with their entity-consistent predecessor and successor context.
The retrieved evidence is serialized and provided with the corresponding event clips to the backbone LVLM, which integrates temporal connections and event-induced state changes to infer cross-event causal relations and generate the answer.
Here, event-causal reasoning denotes question-conditioned inference from observable event evidence: EC-RAG extracts event-local causal transitions and organizes them into entity-consistent temporal trajectories, while the answer model determines the causal direction required by the question. This task differs from structural causal discovery, which aims to identify causal graphs from observational or interventional data.
Importantly, Event-Causal RAG is lightweight and can be integrated with existing video foundation models without expensive end-to-end retraining over ultra-long sequences.

We evaluate our approach on long-video understanding benchmarks. The results show that Event-Causal RAG consistently improves direct inference across multiple video backbones and outperforms the clip-based Video-RAG baseline on completed comparisons, especially on questions that require multi-event integration and causal inference across extended temporal gaps.
Moreover, our method reduces memory usage while maintaining stable performance in streaming scenarios.
In summary, EC-RAG addresses the three core challenges identified above: \textbf{(a)} It mitigates spatiotemporal fragmentation by constructing event-level memory, \textbf{(b)} strengthens long-horizon reasoning by linking event-local causal transitions through entity-centered temporal topology, and \textbf{(c)} reduces storage and online inference overhead through compact dual-store retrieval. These properties make EC-RAG suitable for ultra-long and streaming video scenarios where robust reasoning requires both efficient memory management and temporally grounded causal evidence.

\section{Related Work}
\label{sec:related}
\paragraph{Long Video Understanding via LVLMs.}
Recent large vision-language models (LVLMs) have substantially advanced video understanding, enabling action recognition, temporal description, and question answering over short- and medium-length inputs~\cite{li2024llavanext,he2024malmm,ranasinghe2024understanding,huang2024longvlm}. 
To scale LVLMs to longer videos, existing studies explore frame sampling, temporal pooling, visual token compression, context-window extension, positional encoding adaptation, KV-cache compression, and streaming inference~\cite{wei2024visual,ge2025v2pe,ge2024adaptive,zhao2024minicache,shin2025infinipot,chen2024streaming,anonymous2024streamingvlm,wang2025videoxl}. 
These methods improve the computational feasibility of processing long visual sequences.
However, merely extending the accessible context does not resolve the core challenge of long-video understanding, since the context window of LLMs is ultimately bounded and cannot scale to ultra-long video inputs.
Retrieval-augmented generation (RAG) with external memory offers an alternative by decoupling video processing from question answering~\cite{he2024malmm,jeong2025videorag,luo2024video,VideoRAG}. 
Recent scene-level and graph-based multimodal RAG variants further show that structured memory can improve multi-hop retrieval and long-range reasoning~\cite{wang2024megarag,scenerag2025,vgent2025,vigrag2026}. 
Nevertheless, most existing RAG pipelines still construct memory over fixed-duration clips, coarse scenes, or generic semantic graphs. 
Such memory units can break semantically complete events, while embedding-based retrieval mainly captures query similarity and lacks explicit modeling of event-induced state changes and their long-range temporal connections.

\paragraph{Temporal-Causal Reasoning in Long Videos.}
Prior research on video understanding has extensively studied temporal reasoning, including event ordering, temporal localization, and long-range dependency modeling~\cite{xiao2021next,du2024eventoriented,mangalam2024egoschema,fu2024videomme,liu2024longvideobench,chen2024mecd,liu2024causality}.
Early approaches mainly focus on short clips or temporally localized events, where temporal dynamics can be captured within limited contexts.
To handle longer videos, subsequent works explore hierarchical representations, memory-based architectures, and retrieval-based frameworks to improve the modeling of extended temporal structure~\cite{mangalam2024egoschema,fu2024videomme,liu2024longvideobench}.
Meanwhile, another line of work investigates causal reasoning in videos, such as event causality modeling, causal temporal grounding, and counterfactual analysis~\cite{du2024eventoriented,chen2024mecd,wang2025temporalcot,fei2024video,liu2024causality}, aiming to move beyond temporal correlation toward a deeper understanding of event relations. 
However, these methods are not designed to organize event-local state transitions and entity-consistent trajectories over hour-scale streams.
Unlike prior approaches, EC-RAG represents streaming videos as a global event-state graph. 
This design preserves event-local causal transitions and links them into entity-consistent temporal trajectories, enabling long-range evidence retrieval for ultra-long or streaming video reasoning.

\section{Methodology}\label{sec:method}
\begin{figure*}[t]
    \centering
    \includegraphics[width=\textwidth]{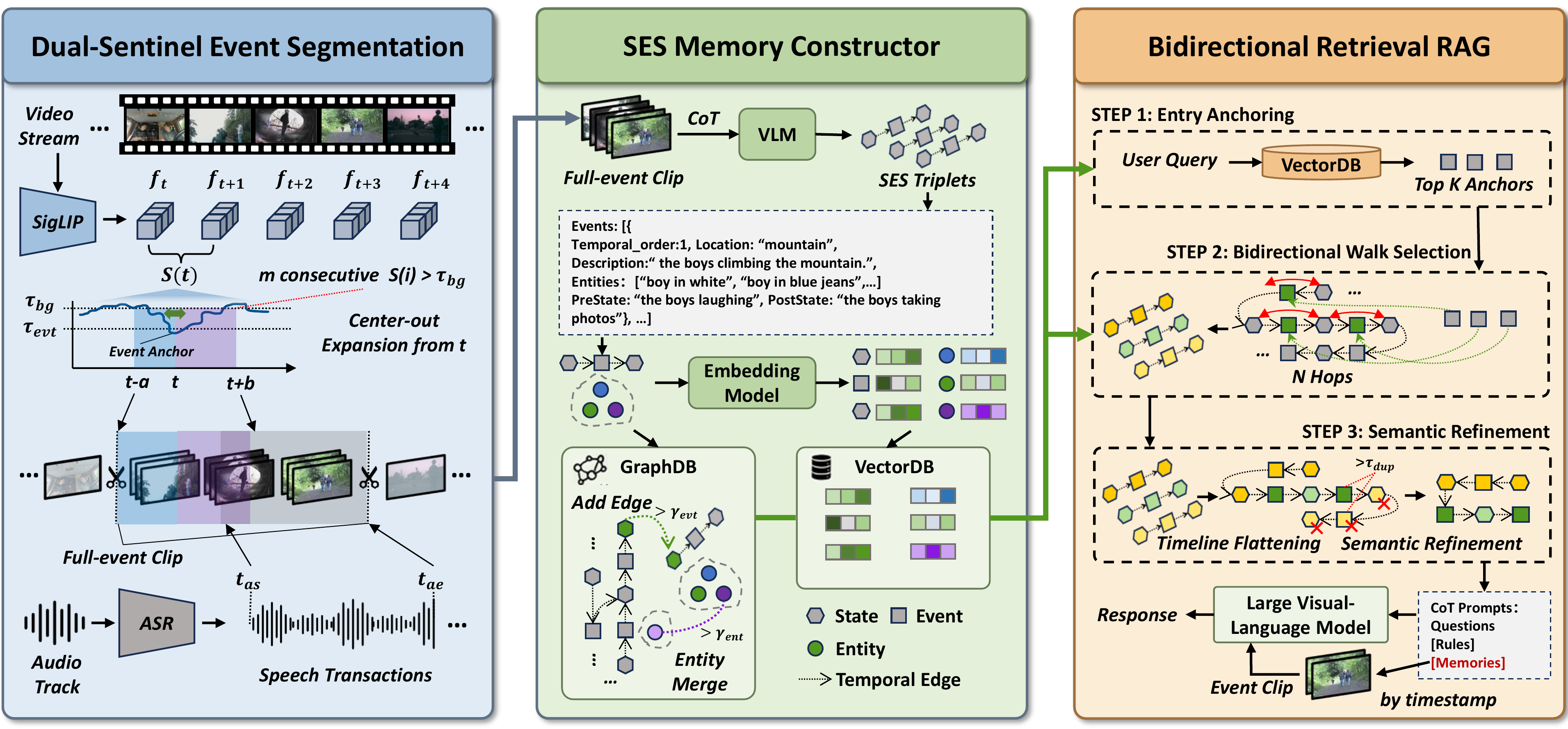}
    \caption{Detailed architecture of EC-RAG. Dual visual-audio sentinels produce event-aligned clips; the SES memory constructor captures event-local state transitions, aligns entities, and builds vector and graph memory; bidirectional retrieval recovers predecessor and successor evidence for the backbone LVLM.}
    \label{fig:overview}
\end{figure*}

Figure~\ref{fig:overview} illustrates EC-RAG. The framework first combines visual changes with timestamped ASR to convert a video stream into event-centric units. Each event is represented as an observed event-local causal transition in State--Event--State (SES) form, and records referring to the same entity are connected into temporal trajectories in dual-store memory. Relevant predecessor and successor evidence is then retrieved to support question-conditioned causal reasoning by the LVLM.

\subsection{Dual-Sentinel Event Segmentation}\label{method:3-1}
\paragraph{Visual Sentinel.}
Let the sampled video stream be $V=\{f_t\}_{t=1}^{T}$, where $f_t$ is the $t$-th frame and $T$ is the number of sampled frames. A pre-trained SigLIP encoder~\cite{zhou2023siglip} $E_v$ maps each frame to an embedding. We measure the visual change between adjacent frames by
\begin{equation}
    S(t)=\cos\!\left(E_v(f_t),E_v(f_{t+1})\right).
\end{equation}
A lower $S(t)$ indicates a larger visual change. In a single scan, we mark $t$ as an event anchor when $S(t)<\tau_{\mathrm{evt}}$ and it is lower than both neighboring values. The local-minimum condition suppresses repeated triggers from frame-level noise. From each anchor, we expand in both directions until $m$ consecutive frame pairs satisfy $S(t)\geq\tau_{\mathrm{bg}}$, where $\tau_{\mathrm{evt}}$ and $\tau_{\mathrm{bg}}$ are the event-trigger and background-stability thresholds, respectively. This produces initial intervals $I_i=[s_i,e_i]$.

Adjacent intervals are merged when $s_{i+1}-e_i<p_{\mathrm{bg}}$, where $p_{\mathrm{bg}}$ is the minimum independent-background duration, preventing short pauses from splitting an action. Events exceeding the maximum unit duration $L_{\max}$ use overlapping $L_{\max}$-bounded units under a single memory identity, preserving boundary context.

\paragraph{Audio Sentinel.}
Visual changes alone can miss speech-driven events and may place a boundary inside an unfinished utterance. We use ASR timestamps and preserved punctuation to group speech into sentence records $(x_j,A_j)$, where $x_j$ is the transcript and $A_j=[t_{as_j},t_{ae_j}]$ is its temporal interval. If $A_j$ overlaps one or more visual events, their boundaries are extended or joined to cover the complete sentence. If no overlap exists, $A_j$ creates an audio-triggered event containing the corresponding video frames. In both cases, $x_j$ is attached to the resulting interval and remains available to SES extraction, retrieval, and answer generation.

After audio-visual fusion, all time ranges outside the event intervals form background segments. Backgrounds are retained rather than discarded: event segments receive dense visual processing, whereas backgrounds are sampled sparsely to preserve continuous scene context at lower cost.

\subsection{State--Event--State Memory Constructor}\label{method:3-2}
\paragraph{Event-First SES Extraction.}
For each event interval, the VLM jointly observes the video segment and its aligned ASR. It identifies the central event and enumerates the observable entities participating in it to ensure coverage of people, objects, and scenes that may otherwise be overlooked. Using the event as the anchor, it then describes each persistent entity's state before the event and its resulting state afterward. This procedure defines SES as an event-centered observed state-transition representation:
\begin{equation}
    \begin{aligned}
    State_i^{\mathrm{pre}}
    &\xrightarrow{\mathrm{PRECONDITION\_OF}} Event_i,\\
    Event_i
    &\xrightarrow{\mathrm{RESULTS\_IN}} State_i^{\mathrm{post}}.
    \end{aligned}
\end{equation}
Here, $State_i^{\mathrm{pre}}$ and $State_i^{\mathrm{post}}$ are the observable states before and after $Event_i$, with the central event anchoring the transition between them within the event interval. The labels \texttt{PRECONDITION\_OF} and \texttt{RESULTS\_IN} encode the event-local causal direction from the observed precondition, through the event, to the resulting state. Each SES record retains its source interval, grounding subsequent graph construction in physical time. The aligned ASR complements visual evidence with spoken interactions and other semantics that may not be visually explicit.

\paragraph{Entity Coreference Alignment.}
The same entity may be described differently across events. Let $m_i$ be a newly extracted entity mention, $\mathcal E$ the set of existing canonical entities, and $\Psi_e$ the text embedding model. We identify its closest existing entity as
\begin{equation}
    e^{*}=\arg\max_{e\in\mathcal E}
    \cos\!\left(\Psi_e(m_i),\Psi_e(e)\right).
\end{equation}
If the maximum similarity is at least $\gamma_{\mathrm{ent}}$, where $\gamma_{\mathrm{ent}}$ is the entity-alignment threshold, $m_i$ is aligned with $e^{*}$; otherwise, it is added to $\mathcal E$ as a new canonical entity. This alignment provides consistent references to people, objects, and scenes across distant events.

\paragraph{Dual-Store Event Memory.}
EC-RAG maintains a vector database for semantic event recall and a graph database for temporal and state-transition topology. The vector store indexes event-level SES records. Through canonical entity references, the graph store organizes local causal transitions into entity-consistent temporal trajectories, preserving successive states, associated events, event order, and directed state continuity.

For two events that share a canonical entity and satisfy $t_i^{e}\leq t_j^{s}$, where $t_i^{e}$ is the end time of $Event_i$ and $t_j^{s}$ is the start time of $Event_j$, we measure state continuity as
\begin{equation}
    r_{ij}=\cos\!\left(
    \Psi_e(State_i^{\mathrm{post}}),
    \Psi_e(State_j^{\mathrm{pre}})
    \right).
\end{equation}
If $r_{ij}\geq\gamma_{\mathrm{evt}}$, with $\gamma_{\mathrm{evt}}$ denoting the event-link threshold, we add the directed edge
\begin{equation}
    Event_i\xrightarrow{\mathrm{TEMPORAL\_NEXT}}Event_j.
\end{equation}
The pre- and post-state representations remain attached to their respective events, while their similarity determines whether the corresponding event nodes should be connected. The resulting \texttt{TEMPORAL\_NEXT} links organize event-local causal transitions into entity-centered spatiotemporal trajectories. The graph therefore preserves how each entity evolves over time, the event associated with each observed state transition, and the temporal connections among those transitions.

\begin{table*}[!t]
  \centering
  \small
  \setlength{\tabcolsep}{4pt}
  \begin{tabular*}{\textwidth}{@{\extracolsep{\fill}}llcccc}
    \toprule
    \textbf{Method} & \textbf{Frames} & \textbf{Forward (554)} & \textbf{Backward (490)} & \textbf{Global (207)} & \textbf{Overall (1,251)} \\
    \midrule
    GPT-5.6 Luna (direct, closed-source) & 96 & 48.92 & 73.67 & 69.57 & \textbf{62.03} \\
    \midrule
    InternVL3.5-8B & 96 & 51.62 & 51.02 & 48.79 & 50.92 \\
    \textbf{+Video-RAG}~\cite{luo2024video} &  & 55.42 & 50.00 & 54.11 & 53.08 \\
    \textbf{+EC-RAG} &  & \textbf{56.86 (+5.23)} & \textbf{54.08 (+3.06)} & \textbf{57.49 (+8.70)} & \textbf{55.88 (+4.96)} \\
    \addlinespace
    VideoLLaMA3-7B & 96 & 50.18 & 41.02 & 48.79 & 46.36 \\
    \textbf{+Video-RAG} &  & -- & -- & -- & -- \\
    \textbf{+EC-RAG} &  & \textbf{59.75 (+9.57)} & \textbf{56.73 (+15.71)} & \textbf{56.52 (+7.73)} & \textbf{58.03 (+11.67)} \\
    \addlinespace
    Qwen3-VL-8B & 96 & 56.32 & 50.20 & 56.04 & 53.88 \\
    \textbf{+Video-RAG} &  & 57.76 & 54.69 & 60.87 & 57.07 \\
    \textbf{+EC-RAG} &  & \textbf{62.09 (+5.78)} & \textbf{58.57 (+8.37)} & \textbf{62.80 (+6.76)} & \textbf{60.83 (+6.95)} \\
    \bottomrule
  \end{tabular*}
  \caption{Accuracy (\%) on ECV-1H with a 96-frame QA budget. Parentheses show absolute gains over paired direct inference. Dashes indicate that Video-RAG exceeded the backbone's native context limit under the same retrieval configuration.}
  \label{tab:hour_scale_results}
\end{table*}

\subsection{Bidirectional Retrieval RAG}\label{method:3-3}
Given a question $q$, EC-RAG decomposes it into retrieval cues describing the queried entities, events, state changes, and temporal direction. The cues retrieve candidates from the vector store, and a reranker scores them against the complete question. The top $K$ event nodes form the anchor set $\mathcal A$ for graph retrieval.

\paragraph{Bidirectional Topological Expansion.}
From each anchor, we traverse both incoming and outgoing graph edges to recover candidate predecessor and successor evidence. Let $d_G(u,v)$ be the shortest directed hop distance in graph $G$, and define the bidirectional distance as $d_{\leftrightarrow}(u,v)=\min\{d_G(u,v),d_G(v,u)\}$. With maximum traversal depth $N$, the retrieved neighborhood is
\begin{equation}
    \mathcal H_N(\mathcal A)=
    \left\{v\mid
    \min_{a\in\mathcal A}d_{\leftrightarrow}(a,v)\leq N
    \right\}.
\end{equation}
This neighborhood contains both predecessors and successors without enumerating all possible paths in a dense graph.

\paragraph{Temporal Ordering and Semantic Deduplication.}
The retrieved events are ordered by their physical timestamps. Because adjacent event windows retain overlapping context, a continuous action or unchanged state may appear in multiple SES records. We compare each candidate's SES embedding with those already retained and remove it when the maximum cosine similarity reaches the deduplication threshold $\tau_{\mathrm{dup}}$.

Finally, EC-RAG serializes the retained nodes according to their timestamps and directed graph relations. Each evidence item contains the SES transition, its aligned ASR, and the source interval, while the corresponding event clips are retrieved from the video store. The resulting evidence exposes two complementary structures: event-local causal transitions within SES records and entity-consistent temporal trajectories across events. The backbone LVLM jointly reasons over these structures, the aligned ASR, and the source clips to connect distributed observations, assess their event-causal relation in the context of the question, and determine the required causal direction.

\section{Experiments and Analysis}
\label{sec:experiments}

\subsection{Settings}
\paragraph{Evaluation Scope.}
We evaluate EC-RAG along three dimensions. ECV-1H and Video-MME Long~\cite{fu2024videomme} measure directed event-causal reasoning and general long-video understanding, respectively; NExT-QA~\cite{xiao2021next} (5--180s) and Event-Bench~\cite{du2024eventoriented} (60--1800s) test transfer to shorter ranges; and a 24-hour stream with controlled VRAM profiling assesses deployment efficiency. On ECV-1H, we also compare with the NeurIPS 2025 Video-RAG baseline~\cite{luo2024video} and measure retrieved-context length. All paired direct and EC-RAG results share the answer backbone and visual-input budget.
\paragraph{Implementation Details.}
Experimental configurations, prompts, and hyperparameters are provided in the supplementary file. We evaluate Qwen3-VL-8B~\cite{bai2025qwen3vl}, InternVL3.5-8B~\cite{wang2025internvl35}, and VideoLLaMA3-7B~\cite{zhang2025videollama3} as open-source backbones. Closed-source models serve only as references and are excluded from paired gain calculations. GPT-4o and Gemini-1.5 Pro results on Video-MME Long and Event-Bench follow the benchmark reports~\cite{fu2024videomme,du2024eventoriented}; GPT-5.6 Luna~\cite{openai2026gpt56} and the NExT-QA closed-source entries use our QA protocol.

\subsection{Long-Video Evaluation}
\label{sec:curated_benchmark}

\paragraph{ECV-1H Benchmark.}
Existing long-video benchmarks evaluate broad long-context understanding, temporal localization, narrative multi-step reasoning, and causal-chain grounding~\cite{liu2024longvideobench,chandrasegaran2024hourvideo,zou2025hlv1k,yu2025vrbench,zhang2026castbench}. To our knowledge, ECV-1H is the first benchmark dedicated to directed event-causal reasoning in real-world untrimmed videos whose source videos all exceed one hour. It contains 123 videos totaling over 150 hours and 1,251 human-authored, independently verified four-choice QA pairs spanning forward, backward, and global event relations. Its publicly accessible, user-uploaded sources cover gameplay recordings, vlogs, tutorials, documentaries, and livestreams.

More than ten in-house annotators followed a two-stage protocol. For each video, a primary annotator watched the complete recording and authored questions, correct answers, distractors, and supporting temporal evidence. A second annotator independently answered each question and verified the sufficiency of its evidence and the validity of its directed causal relation. Human experts and a large language model provided additional quality control. For controlled evaluation, paired variants use identical videos, prompts, 96-frame QA budgets, and decoding settings, with answer options permuted by a fixed seed before scoring.

\paragraph{Main Results.}
Table~\ref{tab:hour_scale_results} reports the hour-scale results. EC-RAG improves all three open-source backbones in every event category, raising overall accuracy by 4.96 points on InternVL3.5-8B, 11.67 points on VideoLLaMA3-7B, and 6.95 points on Qwen3-VL-8B. The largest gain occurs on VideoLLaMA3-7B, which has the lowest direct accuracy, indicating that explicit event memory can compensate for limited native long-context reasoning. Qwen3-VL-8B with EC-RAG attains 60.83\%, the best result among the evaluated open-source backbones and within 1.20 points of the closed-source GPT-5.6 Luna direct-inference reference. These consistent gains show that the benefit is not model- or query-specific.

\begin{table*}[!t]
  \centering
  \small
  \setlength{\tabcolsep}{4pt}
  \begin{tabular*}{\textwidth}{@{\extracolsep{\fill}}lccccc}
    \toprule
    & \textbf{NExT-QA (5--180s)} & \multicolumn{4}{c}{\textbf{Event-Bench (60--1800s)}} \\
    \cmidrule(r){2-2} \cmidrule(r){3-6}
    \textbf{Method} & \textbf{Accuracy (\%)} & \textbf{Contextual} & \textbf{Episodic} & \textbf{Counter-intuitive} & \textbf{Overall} \\
    \midrule
    GPT-4o (closed-source) & 76.71 & 50.13 & 37.33 & 63.44 & 53.33 \\
    Gemini-1.5 Pro (closed-source) & 75.74 & 32.15 & 38.67 & 52.86 & 43.24 \\
    GPT-5.6 Luna (closed-source) & 81.57 & 61.52 & 61.67 & 77.53 & 65.51 \\
    \midrule
    Qwen3-VL-8B & 73.08 & 53.16 & 45.67 & 59.91 & 52.39 \\
    \textbf{+EC-RAG} & \textbf{75.54 (+2.46)} & \textbf{61.77 (+8.61)} & \textbf{48.33 (+2.66)} & 53.30 & \textbf{55.31 (+2.92)} \\
    \addlinespace
    VideoLLaMA3-7B & 76.69 & 56.71 & 42.00 & 55.95 & 51.74 \\
    \textbf{+EC-RAG} & \textbf{78.48 (+1.79)} & \textbf{61.01 (+4.30)} & \textbf{49.67 (+7.67)} & 55.51 & \textbf{55.97 (+4.23)} \\
    \addlinespace
    InternVL3.5-8B & 77.54 & 56.96 & 45.00 & 60.35 & 53.90 \\
    \textbf{+EC-RAG} & \textbf{80.54 (+3.00)} & \textbf{62.28 (+5.32)} & \textbf{46.67 (+1.67)} & 57.27 & \textbf{55.97 (+2.07)} \\
    \bottomrule
  \end{tabular*}
  \caption{Quantitative evaluation on NExT-QA and Event-Bench. Parentheses show absolute gains over the corresponding direct-inference baseline. GPT-4o and Gemini-1.5 Pro Event-Bench scores are reported by the benchmark authors~\cite{du2024eventoriented}.}
  \label{tab:main_results}
\end{table*}

\paragraph{Comparison with Video-RAG and Context Efficiency.}
With the same 96-frame QA budget, EC-RAG outperforms Video-RAG~\cite{luo2024video} on both backbones for which Video-RAG completed inference. On InternVL3.5-8B, EC-RAG improves overall accuracy from 53.08\% to 55.88\% (+2.80 points), with gains of 1.44, 4.08, and 3.38 points on the forward, backward, and global subsets. On Qwen3-VL-8B, it improves overall accuracy from 57.07\% to 60.83\% (+3.76 points), with corresponding gains of 4.33, 3.88, and 1.93 points. Gains in both directed categories are consistent with recovering predecessor and successor evidence rather than relying only on query--clip similarity. Table~\ref{tab:ecv_context_length} shows that EC-RAG reduces average retrieved text from 15,429 to 361 tokens (a 97.66\% reduction, equivalent to a 42.7:1 compression ratio) and the maximum from 52,417 to 696 tokens. Video-RAG cannot produce valid answers with VideoLLaMA3-7B under this configuration because the retrieved context exceeds its native limit. EC-RAG thus provides more compact evidence while improving accuracy on both completed backbone comparisons.

\begin{table}[!t]
  \centering
  \small
  \setlength{\tabcolsep}{2pt}
  \begin{tabular*}{\columnwidth}{@{\extracolsep{\fill}}lrrrr}
    \toprule
    \textbf{Method} & \textbf{Avg. chars} & \textbf{Avg. tokens} & \textbf{Median} & \textbf{Max} \\
    \midrule
    Video-RAG & 22,788 & 15,429 & 12,815 & 52,417 \\
    EC-RAG & \textbf{1,066} & \textbf{361} & \textbf{351} & \textbf{696} \\
    \bottomrule
  \end{tabular*}
  \caption{Retrieved-context length per question on ECV-1H, excluding visual tokens.}
  \label{tab:ecv_context_length}
\end{table}

\paragraph{Generalization to Video-MME Long.}
We further evaluate EC-RAG on the complete 900-question Video-MME Long split. Table~\ref{tab:videomme_results} shows gains of 4.67, 2.78, and 2.67 percentage points for Qwen3-VL-8B, VideoLLaMA3-7B, and InternVL3.5-8B, respectively. Although Video-MME Long covers broader question types rather than specifically targeting directed event causality, these gains show that event-centric memory transfers beyond ECV-1H to general long-video understanding.

\begin{table}[!t]
  \centering
  \small
  \setlength{\tabcolsep}{2.5pt}
  \begin{tabular*}{\columnwidth}{@{\extracolsep{\fill}}lrrrr}
    \toprule
    \textbf{Method} & \textbf{Input} & \textbf{Direct} & \textbf{+EC-RAG} & \textbf{$\Delta$} \\
    \midrule
    GPT-4o (reported) & 384 & 65.30 & -- & -- \\
    Gemini-1.5 Pro (reported) & 1fps & 67.40 & -- & -- \\
    GPT-5.6 Luna (closed-source) & 96 & 61.44 & -- & -- \\
    \midrule
    Qwen3-VL-8B & 96 & 48.33 & \textbf{53.00} & \textbf{+4.67} \\
    VideoLLaMA3-7B & 24 & 42.22 & \textbf{45.00} & \textbf{+2.78} \\
    InternVL3.5-8B & 96 & 52.11 & \textbf{54.78} & \textbf{+2.67} \\
    \bottomrule
  \end{tabular*}
  \caption{Accuracy (\%) on Video-MME Long. Input is the frame count or sampling rate; $\Delta$ is the absolute gain over direct inference. GPT-4o and Gemini-1.5 Pro results are reported by the benchmark authors~\cite{fu2024videomme}; GPT-5.6 Luna follows our QA protocol.}
  \label{tab:videomme_results}
\end{table}

\subsection{Short- and Medium-Video Evaluation}
\paragraph{NExT-QA.}
Table~\ref{tab:main_results} shows that EC-RAG improves all three open-source backbones by 1.79--3.00 percentage points on NExT-QA. Although these videos are substantially shorter than ECV-1H, their questions still require temporal and causal integration across actions. Thus, event-structured retrieval helps even when native context accommodates the video, beyond merely remedying overflow. InternVL3.5-8B with EC-RAG reaches 80.54\%, within 1.03 points of the closed-source GPT-5.6 Luna reference.

\paragraph{Event-Bench.}
EC-RAG improves overall Event-Bench accuracy by 2.07--4.23 points, with the largest gains on \textit{Contextual} (+4.30 to +8.61) and \textit{Episodic} (+1.67 to +7.67) reasoning, where evidence spans events. \textit{Counter-intuitive} accuracy decreases by 0.44--6.61 points. Questions such as ``Why is this video so magical?'' require contrasting abnormal or humorous details with everyday expectations; fact-grounded SES evidence may over-anchor compact backbones to literal descriptions and underweight incongruity cues. Concurrent gains on the other subsets suggest an evidence--task mismatch, not general segmentation noise. Query-adaptive fusion of direct and retrieved predictions may mitigate this trade-off. Overall accuracy still improves for every backbone.

\begin{figure*}[!t]
    \centering
    \includegraphics[width=\textwidth]{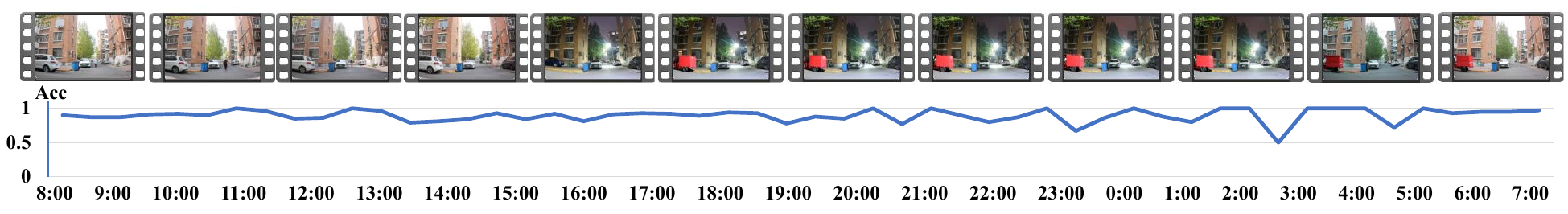}
    \caption{Hourly pass rate under the strict event-validity criteria during the continuous 24-hour surveillance stress test.}
    \label{fig:24hour}
\end{figure*}

\subsection{Ablation Study}
We conduct a controlled text-only ablation on the Event-Bench contextual split using Qwen3-VL-8B (Table~\ref{tab:ablation}) to isolate each RAG component. All variants remove video frames and answer solely from retrieved evidence under fixed retrieval and decoding settings. The complete EC-RAG result, 49.37\%, is the reference for each structural ablation. This isolates RAG structure from repeated visual observations in overlapping event windows.

\begin{table}[!t]
  \centering
  \small
  \setlength{\tabcolsep}{3pt}
  \begin{tabular}{p{0.69\columnwidth}r}
    \toprule
    \textbf{Model Variant} & \textbf{Accuracy (\%)} \\
    \midrule
    \textbf{Complete EC-RAG (Text-Only)} & \textbf{49.37} \\
    \midrule
    $w/o$ Dual-Sentinel event segmentation \newline \textit{(fixed-duration chunks)} & 46.84 (-2.53) \\
    \addlinespace
    $w/o$ SES representation \newline \textit{(uses unstructured event summaries)} & 43.04 (-6.33) \\
    \addlinespace
    $w/o$ Graph topology and bidirectional traversal \newline \textit{(single-turn dense retrieval \cite{luo2024video})} & 38.99 (-10.38) \\
    \addlinespace
    $w/o$ Semantic deduplication \newline \textit{(retains all hop-retrieved evidence)} & 46.33 (-3.04) \\
    \bottomrule
  \end{tabular}
  \caption{Text-only RAG ablation on Event-Bench Contextual. Parentheses show absolute changes from the complete architecture.}
  \label{tab:ablation}
\end{table}

Replacing event-aligned segmentation with fixed-duration chunks reduces accuracy by 2.53 points. Removing SES representation, graph topology and bidirectional traversal, or semantic deduplication reduces accuracy by 6.33, 10.38, and 3.04 points, respectively. These results isolate the contributions of event-aligned segmentation, explicit state transitions, graph-based evidence expansion, and context deduplication in the text-only setting.

\subsection{24-Hour Streaming Deployment Stress Test}
\paragraph{Scope and Protocol.}
This experiment measures system-level streaming stability and throughput; cross-backbone QA accuracy is evaluated separately. We process a continuous, unedited 24-hour industrial surveillance stream on one RTX 5090, using the same Qwen3.5 backbone~\cite{qwen2026qwen35} throughout to generate semantic records from automatically segmented events. Human review accepts an event only if its action, temporal boundary, entity attributes, and location are all correct, yielding an all-or-nothing strict event-validity score.

\paragraph{Results and Interpretation.}
EC-RAG extracts 2,514 events and processes the full stream in 11 hours and 14 minutes, corresponding to approximately 2.14$\times$ real-time throughput. Human review accepts 2,277 events, producing a strict event-validity score of \textbf{90.57\%}; the hourly pass rate remains stable throughout the stream (Figure~\ref{fig:24hour}). Completing the day-long stream verifies continuous event extraction at this duration. Event records remain available after visual-state release, consistent with the bounded residency measured in the VRAM experiment.



\subsection{VRAM Efficiency on Consumer GPU}
\label{sec:vram_analysis}

We profile native Qwen3-VL-8B and EC-RAG on one 32-GB NVIDIA RTX 5090 under matched 2-fps sampling, a 409{,}920-pixel frame budget, SDPA attention, and visual backbone. Native processing accumulates visual inputs; EC-RAG instead uses $L_{\max}=12~\mathrm{s}$ units and releases their visual states while retaining event records.

\begin{figure}[!t]
  \centering
  \includegraphics[width=\columnwidth]{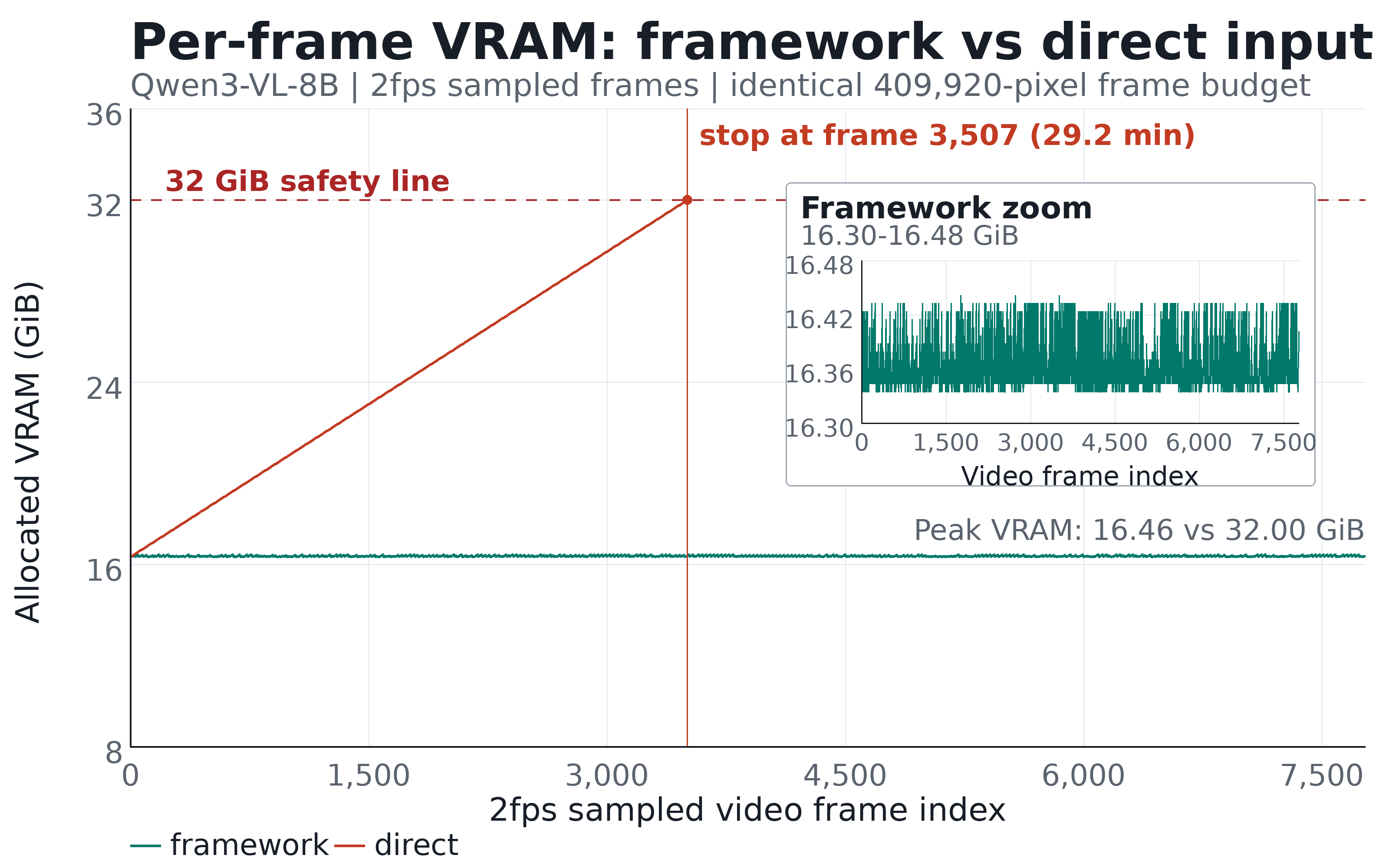}
  \caption{VRAM residency on an RTX 5090 under matched Qwen3-VL-8B settings. Native processing reaches 32~GiB at 3{,}507 frames; EC-RAG completes 3{,}887.2~s with 528 event-level releases while remaining near 16.46~GiB.}
  \label{fig:vram_residency}
\end{figure}

Figure~\ref{fig:vram_residency} shows that native allocation grows almost perfectly linearly ($R^2 \approx 1.0$), by approximately 4.47~GiB per 1{,}000 sampled frames, until reaching 32~GiB at frame 3{,}507. EC-RAG completes the full trace with 528 releases while staying within 16.33--16.46~GiB (17.65~GiB peak device use), preserving approximately 15.5~GiB of headroom and remaining 48.6\% below the native boundary. Its flat curve shows that visual residency depends on the current event unit rather than stream duration, converting duration-dependent accumulation into bounded continuous processing.

\section{Conclusion}

We introduced EC-RAG, a lightweight event-centric retrieval framework for reasoning over ultra-long and streaming video, together with ECV-1H, an hour-scale benchmark for event-causal reasoning over untrimmed videos. EC-RAG combines dual-sentinel segmentation, State–Event–State representations, and dual vector-graph memory to retrieve compact, temporally connected evidence for the answer backbone. Across ECV-1H and three public benchmarks, EC-RAG consistently improves multiple open-source backbones while producing markedly smaller retrieval context than Video-RAG; streaming tests and VRAM profiling further confirm continuous memory updates and bounded visual residency on consumer hardware. Future work will extend ECV-1H and tighten the coupling between segmentation, memory, and adaptive retrieval for evolving streams.

{
\small
\bibliography{refs}

@inproceedings{ranasinghe2024understanding,
  title={Understanding Long Videos with Multimodal Language Models},
  author={Ranasinghe, Kanchana and Li, Xiang and Kahatapitiya, Kumara and Ryoo, Michael S.},
  booktitle={International Conference on Learning Representations (ICLR)},
  year={2025},
  eprint={2403.16998},
  archivePrefix={arXiv},
  primaryClass={cs.CV},
  doi={10.48550/arXiv.2403.16998},
  url={https://arxiv.org/abs/2403.16998}
}

@article{souza2024temporal,
  title={Temporal Contrastive Learning for Video Temporal Reasoning in Large Vision-Language Models},
  author={Souza, Rafael and Lim, Jia-Hao and Davis, Alexander},
  journal={arXiv preprint arXiv:2412.11391},
  year={2024},
  eprint={2412.11391},
  archivePrefix={arXiv},
  primaryClass={cs.CV},
  doi={10.48550/arXiv.2412.11391},
  url={https://arxiv.org/abs/2412.11391}
}

@article{wei2024visual,
  title={Visual Context Window Extension: A New Perspective for Long Video Understanding},
  author={Wei, Hongchen and Chen, Zhenzhong},
  journal={arXiv preprint arXiv:2409.20018},
  year={2024},
  eprint={2409.20018},
  archivePrefix={arXiv},
  primaryClass={cs.CV},
  doi={10.48550/arXiv.2409.20018},
  url={https://arxiv.org/abs/2409.20018}
}

@inproceedings{ge2025v2pe,
  title={{V2PE}: Improving Multimodal Long-Context Capability of Vision-Language Models with Variable Visual Position Encoding},
  author={Ge, Junqi and Chen, Ziyi and Lin, Jintao and Zhu, Jinguo and Liu, Xihui and Dai, Jifeng and Zhu, Xizhou},
  booktitle={Proceedings of the IEEE/CVF International Conference on Computer Vision (ICCV)},
  pages={21070--21084},
  year={2025},
  eprint={2412.09616},
  archivePrefix={arXiv},
  primaryClass={cs.CV},
  url={https://openaccess.thecvf.com/content/ICCV2025/html/Ge_V2PE_Improving_Multimodal_Long-Context_Capability_of_Vision-Language_Models_with_Variable_ICCV_2025_paper.html}
}

@article{zhang2024survey,
  title={From Seconds to Hours: Reviewing MultiModal Large Language Models on Comprehensive Long Video Understanding},
  author={Zou, Heqing and Luo, Tianze and Xie, Guiyang and Zhang, Victor and Lv, Fengmao and Wang, Guangcong and Chen, Junyang and Wang, Zhuochen and Zhang, Hansheng and Zhang, Huaijian},
  journal={arXiv preprint arXiv:2409.18938},
  year={2024},
  eprint={2409.18938},
  archivePrefix={arXiv},
  primaryClass={cs.CV},
  doi={10.48550/arXiv.2409.18938},
  url={https://arxiv.org/abs/2409.18938}
}

@inproceedings{liu2024longvideobench,
  title={{LongVideoBench}: A Benchmark for Long-context Interleaved Video-Language Understanding},
  author={Wu, Haoning and Li, Dongxu and Chen, Bei and Li, Junnan},
  booktitle={Advances in Neural Information Processing Systems 37 (NeurIPS 2024), Datasets and Benchmarks Track},
  year={2024},
  doi={10.52202/079017-0907},
  eprint={2407.15754},
  archivePrefix={arXiv},
  primaryClass={cs.CV},
  url={https://proceedings.neurips.cc/paper_files/paper/2024/hash/329ad516cf7a6ac306f29882e9c77558-Abstract-Datasets_and_Benchmarks_Track.html}
}

@inproceedings{chen2024streaming,
  title={Streaming Long Video Understanding with Large Language Models},
  author={Qian, Rui and Dong, Xiaoyi and Zhang, Pan and Zang, Yuhang and Ding, Shuangrui and Lin, Dahua and Wang, Jiaqi},
  booktitle={Advances in Neural Information Processing Systems 37 (NeurIPS 2024), Main Conference Track},
  year={2024},
  doi={10.52202/079017-3792},
  eprint={2405.16009},
  archivePrefix={arXiv},
  primaryClass={cs.CV},
  url={https://proceedings.neurips.cc/paper_files/paper/2024/hash/d7ce06e9293c3d8e6cb3f80b4157f875-Abstract-Conference.html}
}

@inproceedings{jeong2025videorag,
  title={{VideoRAG}: Retrieval-Augmented Generation over Video Corpus},
  author={Jeong, Soyeong and Kim, Kangsan and Baek, Jinheon and Hwang, Sung Ju},
  booktitle={Findings of the Association for Computational Linguistics: ACL 2025},
  pages={21278--21298},
  publisher={Association for Computational Linguistics},
  address={Vienna, Austria},
  year={2025},
  doi={10.18653/v1/2025.findings-acl.1096},
  eprint={2501.05874},
  archivePrefix={arXiv},
  url={https://aclanthology.org/2025.findings-acl.1096/}
}

@inproceedings{luo2024video,
  title={{Video-RAG}: Visually-aligned Retrieval-Augmented Long Video Comprehension},
  author={Luo, Yongdong and Zheng, Xiawu and Li, Guilin and Yin, Shukang and Lin, Haojia and Fu, Chaoyou and Huang, Jinfa and Ji, Jiayi and Chao, Fei and Luo, Jiebo and Ji, Rongrong},
  booktitle={Advances in Neural Information Processing Systems (NeurIPS)},
  year={2025},
  eprint={2411.13093},
  archivePrefix={arXiv},
  primaryClass={cs.CV},
  url={https://openreview.net/forum?id=QaZxGWlbgO}
}

@inproceedings{wang2024megarag,
  title={{MegaRAG}: Multimodal Knowledge Graph-Based Retrieval Augmented Generation},
  author={Hsiao, Chi-Hsiang and Wang, Yi-Cheng and Lin, Tzung-Sheng and Yeh, Yi-Ren and Chen, Chu-Song},
  booktitle={Proceedings of the 64th Annual Meeting of the Association for Computational Linguistics (Volume 1: Long Papers)},
  pages={48031--48059},
  address={San Diego, California, United States},
  publisher={Association for Computational Linguistics},
  year={2026},
  doi={10.18653/v1/2026.acl-long.2218},
  url={https://aclanthology.org/2026.acl-long.2218/}
}

@article{VideoRAG,
  title={VideoRAG: Retrieval-Augmented Generation with Extreme Long-Context Videos},
  author={Ren, Xubin and Xu, Lingrui and Xia, Long and Wang, Shuaiqiang and Yin, Dawei and Huang, Chao},
  journal={arXiv preprint arXiv:2502.01549},
  year={2025}
}

@inproceedings{xiao2021next,
  title={{NExT-QA}: Next Phase of Question-Answering to Explaining Temporal Actions},
  author={Xiao, Junbin and Shang, Xindi and Yao, Angela and Chua, Tat-Seng},
  booktitle={Proceedings of the IEEE/CVF Conference on Computer Vision and Pattern Recognition (CVPR)},
  year={2021},
  doi={10.1109/CVPR46437.2021.00965},
  eprint={2105.08276},
  archivePrefix={arXiv},
  primaryClass={cs.CV},
  url={https://arxiv.org/abs/2105.08276}
}

@misc{du2024eventoriented,
    title={Towards Event-oriented Long Video Understanding},
    author={Yifan Du and Kun Zhou and Yuqi Huo and Yifan Li and Wayne Xin Zhao and Haoyu Lu and Zijia Zhao and Bingning Wang and Weipeng Chen and Ji-Rong Wen},
    year={2024},
    eprint={2406.14129},
    archivePrefix={arXiv},
    primaryClass={cs.CV}
}

@inproceedings{mangalam2024egoschema,
  title={{EgoSchema}: A Diagnostic Benchmark for Very Long-form Video Language Understanding},
  author={Mangalam, Karttikeya and Akshulakov, Raiymbek and Malik, Jitendra},
  booktitle={Advances in Neural Information Processing Systems 36 (NeurIPS 2023), Datasets and Benchmarks Track},
  year={2023},
  eprint={2308.09126},
  archivePrefix={arXiv},
  primaryClass={cs.CV},
  url={https://papers.nips.cc/paper_files/paper/2023/hash/90ce332aff156b910b002ce4e6880dec-Abstract-Datasets_and_Benchmarks.html}
}

@inproceedings{fu2024videomme,
  title={{Video-MME}: The First-Ever Comprehensive Evaluation Benchmark of Multi-modal LLMs in Video Analysis},
  author={Fu, Chaoyou and Dai, Yuhan and Luo, Yongdong and Li, Lei and Ren, Shuhuai and Zhang, Renrui and Wang, Zihan and Zhou, Chenyu and Shen, Yunhang and Zhang, Mengdan and Chen, Peixian and Li, Yanwei and Lin, Shaohui and Zhao, Sirui and Li, Ke and Xu, Tong and Zheng, Xiawu and Chen, Enhong and Shan, Caifeng and He, Ran and Sun, Xing},
  booktitle={Proceedings of the IEEE/CVF Conference on Computer Vision and Pattern Recognition (CVPR)},
  pages={24108--24118},
  year={2025},
  eprint={2405.21075},
  archivePrefix={arXiv},
  primaryClass={cs.CV},
  url={https://openaccess.thecvf.com/content/CVPR2025/html/Fu_Video-MME_The_First-Ever_Comprehensive_Evaluation_Benchmark_of_Multi-modal_LLMs_in_CVPR_2025_paper.html}
}

@inproceedings{chen2024mecd,
  title={{MECD}: Unlocking Multi-Event Causal Discovery in Video Reasoning},
  author={Chen, Tieyuan and Liu, Huabin and He, Tianyao and Chen, Yihang and Gan, Chaofan and Ma, Xiao and Zhong, Cheng and Zhang, Yang and Wang, Yingxue and Lin, Hui and Lin, Weiyao},
  booktitle={Advances in Neural Information Processing Systems (NeurIPS)},
  year={2024},
  note={Spotlight paper},
  eprint={2409.17647},
  archivePrefix={arXiv},
  primaryClass={cs.CV},
  doi={10.48550/arXiv.2409.17647},
  url={https://arxiv.org/abs/2409.17647}
}

@inproceedings{rodin2024action,
  title={Action Scene Graphs for Long-Form Understanding of Egocentric Videos},
  author={Rodin, Ivan and Furnari, Antonino and Min, Kyle and Tripathi, Subarna and Farinella, Giovanni Maria},
  booktitle={Proceedings of the IEEE/CVF Conference on Computer Vision and Pattern Recognition (CVPR)},
  pages={18622--18632},
  year={2024},
  url={https://openaccess.thecvf.com/content/CVPR2024/html/Rodin_Action_Scene_Graphs_for_Long-Form_Understanding_of_Egocentric_Videos_CVPR_2024_paper.html}
}

@article{zhang2024multimodal,
  title={Multimodal Event Causality Reasoning with Scene Graph Enhanced Interaction Network},
  author={Liu, Jintao and Wei, Kaiwen and Liu, Chenglong},
  journal={Proceedings of the AAAI Conference on Artificial Intelligence},
  volume={38},
  number={8},
  pages={8778--8786},
  year={2024},
  doi={10.1609/aaai.v38i8.28724},
  url={https://ojs.aaai.org/index.php/AAAI/article/view/28724}
}

@article{li2024causaltad,
  title={Harnessing Temporal Causality for Advanced Temporal Action Detection},
  author={Liu, Shuming and Sui, Lin and Zhang, Chen-Lin and Mu, Fangzhou and Zhao, Chen and Ghanem, Bernard},
  journal={arXiv preprint arXiv:2407.17792},
  year={2024},
  eprint={2407.17792},
  archivePrefix={arXiv},
  primaryClass={cs.CV},
  doi={10.48550/arXiv.2407.17792},
  url={https://arxiv.org/abs/2407.17792}
}

@article{wang2025temporalcot,
  title={Temporal Chain of Thought: Long-Video Understanding by Thinking in Frames},
  author={Arnab, Anurag and Iscen, Ahmet and Caron, Mathilde and Fathi, Alireza and Schmid, Cordelia},
  journal={arXiv preprint arXiv:2507.02001},
  year={2025},
  eprint={2507.02001},
  archivePrefix={arXiv},
  primaryClass={cs.CV},
  doi={10.48550/arXiv.2507.02001},
  url={https://arxiv.org/abs/2507.02001}
}

@inproceedings{fei2024video,
  title={Video-of-Thought: Step-by-Step Video Reasoning from Perception to Cognition},
  author={Fei, Hao and Wu, Shengqiong and Ji, Wei and Zhang, Hanwang and Zhang, Meishan and Lee, Mong-Li and Hsu, Wynne},
  booktitle={Proceedings of the 41st International Conference on Machine Learning},
  series={Proceedings of Machine Learning Research},
  volume={235},
  pages={13109--13125},
  year={2024},
  publisher={PMLR},
  url={https://proceedings.mlr.press/v235/fei24a.html}
}

@article{liu2024causality,
  title={Causality Matters: How Temporal Information Emerges in Video Language Models},
  author={Shi, Yumeng and Long, Quanyu and Wu, Yin and Wang, Wenya},
  journal={arXiv preprint arXiv:2508.11576},
  year={2026},
  note={AAAI 2026},
  eprint={2508.11576},
  archivePrefix={arXiv},
  primaryClass={cs.CV},
  url={https://arxiv.org/abs/2508.11576}
}

@article{dai2024corm,
  title={{CORM}: Cache Optimization with Recent Message for Large Language Model Inference},
  author={Dai, Jincheng and Huang, Zhuowei and Jiang, Haiyun and Chen, Chen and Cai, Deng and Bi, Wei and Shi, Shuming},
  journal={arXiv preprint arXiv:2404.15949},
  year={2024},
  eprint={2404.15949},
  archivePrefix={arXiv},
  primaryClass={cs.CL},
  doi={10.48550/arXiv.2404.15949},
  url={https://arxiv.org/abs/2404.15949}
}

@inproceedings{ge2024adaptive,
  title={Model Tells You What to Discard: Adaptive KV Cache Compression for LLMs},
  author={Ge, Suyu and Zhang, Yunan and Liu, Liyuan and Zhang, Minjia and Han, Jiawei and Gao, Jianfeng},
  booktitle={International Conference on Learning Representations (ICLR)},
  year={2024},
  note={Oral},
  eprint={2310.01801},
  archivePrefix={arXiv},
  primaryClass={cs.CL},
  url={https://openreview.net/forum?id=uNrFpDPMyo}
}

@inproceedings{liu2024pyramid,
  title={PyramidInfer: Pyramid KV Cache Compression for High-throughput LLM Inference},
  author={Yang, Dongjie and Han, Xiaodong and Gao, Yan and Hu, Yao and Zhang, Shilin and Zhao, Hai},
  booktitle={Findings of the Association for Computational Linguistics: ACL 2024},
  pages={3258--3270},
  year={2024},
  doi={10.18653/v1/2024.findings-acl.195},
  eprint={2405.12532},
  archivePrefix={arXiv},
  primaryClass={cs.CL},
  url={https://aclanthology.org/2024.findings-acl.195/}
}

@inproceedings{zhao2024minicache,
  title={{MiniCache}: KV Cache Compression in Depth Dimension for Large Language Models},
  author={Liu, Akide and Liu, Jing and Pan, Zizheng and He, Yefei and Haffari, Gholamreza and Zhuang, Bohan},
  booktitle={Advances in Neural Information Processing Systems 37 (NeurIPS 2024), Main Conference Track},
  year={2024},
  doi={10.52202/079017-4443},
  eprint={2405.14366},
  archivePrefix={arXiv},
  primaryClass={cs.LG},
  url={https://proceedings.neurips.cc/paper_files/paper/2024/hash/fd0705710bf01b88a60a3d479ea341d9-Abstract-Conference.html}
}

@inproceedings{shin2025infinipot,
  title={{InfiniPot-V}: Memory-Constrained KV Cache Compression for Streaming Video Understanding},
  author={Kim, Minsoo and Shim, Kyuhong and Choi, Jungwook and Chang, Simyung},
  booktitle={Advances in Neural Information Processing Systems (NeurIPS)},
  year={2025},
  eprint={2506.15745},
  archivePrefix={arXiv},
  primaryClass={eess.IV},
  url={https://openreview.net/forum?id=hFxOZjHyTg}
}

@inproceedings{zheng2024cachegen,
  title={{CacheGen}: KV Cache Compression and Streaming for Fast Large Language Model Serving},
  author={Liu, Yuhan and Li, Hanchen and Cheng, Yihua and Ray, Siddhant and Huang, Yuyang and Zhang, Qizheng and Du, Kuntai and Yao, Jiayi and Lu, Shan and Ananthanarayanan, Ganesh and Maire, Michael and Hoffmann, Henry and Holtzman, Ari and Jiang, Junchen},
  booktitle={Proceedings of the ACM SIGCOMM Conference},
  year={2024},
  doi={10.1145/3651890.3672274},
  eprint={2310.07240},
  archivePrefix={arXiv},
  url={https://dl.acm.org/doi/10.1145/3651890.3672274}
}

@inproceedings{anonymous2024streamingvlm,
  title={{StreamingVLM}: Real-Time Understanding for Infinite Video Streams},
  author={Xu, Ruyi and Xiao, Guangxuan and Chen, Yukang and He, Liuning and Peng, Kelly and Lu, Yao and Han, Song},
  booktitle={International Conference on Learning Representations (ICLR)},
  year={2026},
  eprint={2510.09608},
  archivePrefix={arXiv},
  primaryClass={cs.CV},
  url={https://openreview.net/forum?id=gVbPWbA97s}
}

@inproceedings{yan2025streaming,
  title={Learning Streaming Video Representation via Multitask Training},
  author={Yan, Yibin and Xu, Jilan and Di, Shangzhe and Liu, Yikun and Shi, Yudi and Chen, Qirui and Li, Zeqian and Huang, Yifei and Xie, Weidi},
  booktitle={Proceedings of the IEEE/CVF International Conference on Computer Vision (ICCV)},
  pages={9900--9912},
  year={2025},
  eprint={2504.20041},
  archivePrefix={arXiv},
  primaryClass={cs.CV},
  url={https://openaccess.thecvf.com/content/ICCV2025/html/Yan_Learning_Streaming_Video_Representation_via_Multitask_Training_ICCV_2025_paper.html}
}

@inproceedings{chatterjee2025streamingvllm,
  title={Streaming VideoLLMs for Real-Time Procedural Video Understanding},
  author={Chatterjee, Dibyadip and Remelli, Edoardo and Song, Yale and Tekin, Bugra and Mittal, Abhay and Bhatnagar, Bharat and Camgoz, Necati Cihan and Hampali, Shreyas and Sauser, Eric and Ma, Shugao and Yao, Angela and Sener, Fadime},
  booktitle={Proceedings of the IEEE/CVF International Conference on Computer Vision (ICCV)},
  pages={22586--22598},
  year={2025},
  eprint={2504.13915},
  archivePrefix={arXiv},
  primaryClass={cs.CV},
  url={https://openaccess.thecvf.com/content/ICCV2025/html/Chatterjee_Streaming_VideoLLMs_for_Real-Time_Procedural_Video_Understanding_ICCV_2025_paper.html}
}

@inproceedings{he2024malmm,
  title={{MA-LMM}: Memory-Augmented Large Multimodal Model for Long-Term Video Understanding},
  author={He, Bo and Li, Hengduo and Jang, Young Kyun and Jia, Menglin and Cao, Xuefei and Shah, Ashish and Shrivastava, Abhinav and Lim, Ser-Nam},
  booktitle={Proceedings of the IEEE/CVF Conference on Computer Vision and Pattern Recognition (CVPR)},
  pages={13504--13514},
  year={2024},
  eprint={2404.05726},
  archivePrefix={arXiv},
  primaryClass={cs.CV},
  doi={10.1109/CVPR52733.2024.01282},
  url={https://openaccess.thecvf.com/content/CVPR2024/html/He_MA-LMM_Memory-Augmented_Large_Multimodal_Model_for_Long-Term_Video_Understanding_CVPR_2024_paper.html}
}

@inproceedings{zhao2025streamchat,
  title={Streaming Video Understanding and Multi-round Interaction with Memory-enhanced Knowledge},
  author={Xiong, Haomiao and Yang, Zongxin and Yu, Jiazuo and Zhuge, Yunzhi and Zhang, Lu and Zhu, Jiawen and Lu, Huchuan},
  booktitle={International Conference on Learning Representations (ICLR)},
  year={2025},
  eprint={2501.13468},
  archivePrefix={arXiv},
  primaryClass={cs.CV},
  url={https://openreview.net/forum?id=JbPb6RieNC}
}

@inproceedings{zhou2023siglip,
  title={Sigmoid Loss for Language Image Pre-Training},
  author={Zhai, Xiaohua and Mustafa, Basil and Kolesnikov, Alexander and Beyer, Lucas},
  booktitle={Proceedings of the IEEE/CVF International Conference on Computer Vision (ICCV)},
  year={2023}
}

@article{li2024llavanext,
  title={{LLaVA-NeXT-Interleave}: Tackling Multi-image, Video, and 3D in Large Multimodal Models},
  author={Li, Feng and Zhang, Renrui and Zhang, Hao and Zhang, Yuanhan and Li, Bo and Li, Wei and Ma, Zejun and Li, Chunyuan},
  journal={arXiv preprint arXiv:2407.07895},
  year={2024},
  eprint={2407.07895},
  archivePrefix={arXiv},
  primaryClass={cs.CV},
  doi={10.48550/arXiv.2407.07895},
  url={https://arxiv.org/abs/2407.07895}
}

@inproceedings{huang2024longvlm,
  title={{LongVLM}: Efficient Long Video Understanding via Large Language Models},
  author={Weng, Yuetian and Han, Mingfei and He, Haoyu and Chang, Xiaojun and Zhuang, Bohan},
  booktitle={Proceedings of the European Conference on Computer Vision (ECCV)},
  pages={453--470},
  year={2024},
  doi={10.1007/978-3-031-73414-4_26},
  eprint={2404.03384},
  archivePrefix={arXiv},
  primaryClass={cs.CV},
  url={https://doi.org/10.1007/978-3-031-73414-4_26}
}

@inproceedings{wang2025videoxl,
  title={{Video-XL}: Extra-Long Vision Language Model for Hour-Scale Video Understanding},
  author={Shu, Yan and Liu, Zheng and Zhang, Peitian and Qin, Minghao and Zhou, Junjie and Liang, Zhengyang and Huang, Tiejun and Zhao, Bo},
  booktitle={Proceedings of the IEEE/CVF Conference on Computer Vision and Pattern Recognition (CVPR)},
  pages={26160--26169},
  year={2025},
  doi={10.1109/CVPR52734.2025.02436},
  eprint={2409.14485},
  archivePrefix={arXiv},
  primaryClass={cs.CV},
  url={https://openaccess.thecvf.com/content/CVPR2025/html/Shu_Video-XL_Extra-Long_Vision_Language_Model_for_Hour-Scale_Video_Understanding_CVPR_2025_paper.html}
}

@inproceedings{wang2025eventvad,
  title={{EventVAD}: Training-Free Event-Aware Video Anomaly Detection},
  author={Shao, Yihua and He, Haojin and Li, Sijie and Chen, Siyu and Long, Xinwei and Zeng, Fanhu and Fan, Yuxuan and Zhang, Muyang and Yan, Ziyang and Ma, Ao and Wang, Xiaochen and Tang, Hao and Wang, Yan and Li, Shuyan},
  booktitle={Proceedings of the 33rd ACM International Conference on Multimedia (MM)},
  pages={2586--2595},
  year={2025},
  doi={10.1145/3746027.3754500},
  eprint={2504.13092},
  archivePrefix={arXiv},
  primaryClass={cs.CV},
  url={https://dl.acm.org/doi/10.1145/3746027.3754500}
}

@article{ali2024bilstm,
  title={Deep BiLSTM Attention Model for Spatial and Temporal Anomaly Detection in Video Surveillance},
  author={Natha, Sarfaraz and Ahmed, Fareed and Siraj, Mohammad and Lagari, Mehwish and Altamimi, Majid and Chandio, Asghar Ali},
  journal={Sensors},
  volume={25},
  number={1},
  pages={251},
  publisher={MDPI},
  year={2025},
  doi={10.3390/s25010251},
  url={https://www.mdpi.com/1424-8220/25/1/251}
}

@article{liu2024lost,
  title={Lost in the Middle: How Language Models Use Long Contexts},
  author={Liu, Nelson F. and Lin, Kevin and Hewitt, John and Paranjape, Ashwin and Bevilacqua, Michele and Petroni, Fabio and Liang, Percy},
  journal={Transactions of the Association for Computational Linguistics},
  volume={12},
  pages={157--173},
  year={2024},
  doi={10.1162/tacl_a_00638},
  url={https://aclanthology.org/2024.tacl-1.9/}
}

@Article{s22176563,
AUTHOR = {Khan, Sardar Waqar and Hafeez, Qasim and Khalid, Muhammad Irfan and Alroobaea, Roobaea and Hussain, Saddam and Iqbal, Jawaid and Almotiri, Jasem and Ullah, Syed Sajid},
TITLE = {Anomaly Detection in Traffic Surveillance Videos Using Deep Learning},
JOURNAL = {Sensors},
VOLUME = {22},
YEAR = {2022},
NUMBER = {17},
ARTICLE-NUMBER = {6563},
URL = {https://www.mdpi.com/1424-8220/22/17/6563},
PubMedID = {36081022},
ISSN = {1424-8220},
DOI = {10.3390/s22176563}
}

@article{pawar2022roadaccidents,
  title={Deep learning based detection and localization of road accidents from traffic surveillance videos},
  author={Pawar, Ketan and Attar, Vahida Z.},
  journal={ICT Express},
  volume={8},
  number={3},
  pages={379--387},
  year={2022},
  doi={10.1016/j.icte.2021.11.004},
  url={https://www.sciencedirect.com/science/article/pii/S2405959521001478}
}

@article{scenerag2025,
  title={SceneRAG: Scene-level Retrieval-Augmented Generation for Video Understanding},
  author={Zeng, Nianbo and Hou, Haowen and Yu, Fei Richard and Shi, Si and He, Ying Tiffany},
  journal={arXiv preprint arXiv:2506.07600},
  year={2025},
  eprint={2506.07600},
  archivePrefix={arXiv},
  primaryClass={cs.CV},
  doi={10.48550/arXiv.2506.07600},
  url={https://arxiv.org/abs/2506.07600}
}

@inproceedings{vgent2025,
  title={Vgent: Graph-based Retrieval-Reasoning-Augmented Generation For Long Video Understanding},
  author={Shen, Xiaoqian and Zhang, Wenxuan and Chen, Jun and Elhoseiny, Mohamed},
  booktitle={Advances in Neural Information Processing Systems},
  year={2025},
  note={Spotlight},
  url={https://openreview.net/forum?id=5xPvWat3IX}
}

@inproceedings{vigrag2026,
  title={{ViG-RAG}: Video-aware Graph Retrieval-Augmented Generation via Temporal and Semantic Hybrid Reasoning},
  author={Cao, Zongsheng and Liu, Anran and He, Yangfan and Li, Jing and Zhang, Bo and Wang, Zigan},
  booktitle={Proceedings of the AAAI Conference on Artificial Intelligence},
  volume={40},
  pages={48--56},
  year={2026},
  doi={10.1609/aaai.v40i1.36963},
  url={https://ojs.aaai.org/index.php/AAAI/article/view/36963}
}

@article{chandrasegaran2024hourvideo,
  title={{HourVideo}: 1-Hour Video-Language Understanding},
  author={Chandrasegaran, Keshigeyan and Gupta, Agrim and Hadzic, Lea M. and Kota, Taran and He, Jimming and Eyzaguirre, Cristobal and Durante, Zane and Li, Manling and Wu, Jiajun and Fei-Fei, Li},
  journal={arXiv preprint arXiv:2411.04998},
  year={2024},
  eprint={2411.04998},
  archivePrefix={arXiv},
  primaryClass={cs.CV},
  url={https://arxiv.org/abs/2411.04998}
}

@article{zou2025hlv1k,
  title={{HLV-1K}: A Large-scale Hour-Long Video Benchmark for Time-Specific Long Video Understanding},
  author={Zou, Heqing and Luo, Tianze and Xie, Guiyang and Zhang, Victor and Lv, Fengmao and Wang, Guangcong and Chen, Junyang and Wang, Zhuochen and Zhang, Hansheng and Zhang, Huaijian},
  journal={arXiv preprint arXiv:2501.01645},
  year={2025},
  eprint={2501.01645},
  archivePrefix={arXiv},
  primaryClass={cs.CV},
  url={https://arxiv.org/abs/2501.01645}
}

@inproceedings{yu2025vrbench,
  title={{VRBench}: A Benchmark for Multi-Step Reasoning in Long Narrative Videos},
  author={Yu, Jiashuo and Wu, Yue and Chu, Meng and Ren, Zhifei and Huang, Zizheng and Pei, Chu and Li, Ruijie and He, Yinan and Li, Qirui and Li, Songze and Li, Zhenxiang and Tu, Zhongying and He, Conghui and Qiao, Yu and Wang, Yali and Wang, Yi and Wang, Limin},
  booktitle={Proceedings of the IEEE/CVF International Conference on Computer Vision},
  pages={21655--21666},
  year={2025}
}

@inproceedings{zhang2026castbench,
  title={{CaST-Bench}: Benchmarking Causal Chain-Grounded Spatio-Temporal Reasoning for Video Question Answering},
  author={Zhang, Mingfang and Pan, Jingjing and Kumar, Ashutosh and Saini, Rajat and Erdogan, Mustafa and Yang, Hsuan-Kung and Kang, Caixin and Huang, Yifei and Sato, Yoichi and Kong, Quan},
  booktitle={Proceedings of the IEEE/CVF Conference on Computer Vision and Pattern Recognition},
  pages={31856--31866},
  year={2026}
}

@article{bai2025qwen3vl,
  title={{Qwen3-VL} Technical Report},
  author={Bai, Shuai and Cai, Yuxuan and Chen, Ruizhe and Chen, Keqin and Chen, Xionghui and Cheng, Zesen and Deng, Lianghao and Ding, Wei and Gao, Chang and Ge, Chunjiang and others},
  journal={arXiv preprint arXiv:2511.21631},
  year={2025},
  eprint={2511.21631},
  archivePrefix={arXiv},
  primaryClass={cs.CV},
  url={https://arxiv.org/abs/2511.21631}
}

@article{wang2025internvl35,
  title={{InternVL3.5}: Advancing Open-Source Multimodal Models in Versatility, Reasoning, and Efficiency},
  author={Wang, Weiyun and Gao, Zhangwei and Gu, Lixin and Pu, Hengjun and Cui, Long and Wei, Xingguang and Liu, Zhaoyang and Jing, Linglin and Ye, Shenglong and Shao, Jie and others},
  journal={arXiv preprint arXiv:2508.18265},
  year={2025},
  eprint={2508.18265},
  archivePrefix={arXiv},
  primaryClass={cs.CV},
  url={https://arxiv.org/abs/2508.18265}
}

@article{zhang2025videollama3,
  title={{VideoLLaMA 3}: Frontier Multimodal Foundation Models for Image and Video Understanding},
  author={Zhang, Boqiang and Li, Kehan and Cheng, Zesen and Hu, Zhiqiang and Yuan, Yuqian and Chen, Guanzheng and Leng, Sicong and Jiang, Yuming and Zhang, Hang and Li, Xin and Jin, Peng and Zhang, Wenqi and Wang, Fan and Bing, Lidong and Zhao, Deli},
  journal={arXiv preprint arXiv:2501.13106},
  year={2025},
  eprint={2501.13106},
  archivePrefix={arXiv},
  primaryClass={cs.CV},
  url={https://arxiv.org/abs/2501.13106}
}

@article{sathyanarayana2018vision,
  title={Vision-Based Patient Monitoring: A Comprehensive Review of Algorithms and Technologies},
  author={Sathyanarayana, Supriya and Satzoda, Ravi Kumar and Sathyanarayana, Suchitra and Thambipillai, Srikanthan},
  journal={Journal of Ambient Intelligence and Humanized Computing},
  volume={9},
  number={2},
  pages={225--251},
  year={2018},
  doi={10.1007/s12652-015-0328-1},
  url={https://doi.org/10.1007/s12652-015-0328-1}
}

@misc{qwen2026qwen35,
  author={{Qwen Team}},
  title={{Qwen3.5}: Towards Native Multimodal Agents},
  year={2026},
  howpublished={Official Qwen release report},
  url={https://qwen.ai/blog?id=qwen3.5}
}

@misc{openai2026gpt56,
  author={{OpenAI}},
  title={{GPT-5.6}: Frontier Intelligence That Scales with Your Ambition},
  year={2026},
  howpublished={OpenAI release report},
  url={https://openai.com/index/gpt-5-6/}
}
}
\end{document}